\documentclass[runningheads]{llncs}
\usepackage{graphicx}
\usepackage{amsmath,amssymb} 
\usepackage{color}

\usepackage{import}
\usepackage{url}
\usepackage{hyperref}
\usepackage{times}
\usepackage{epsfig}

\providecommand\onlynonthesis{}
\providecommand\onlythesis{}
\renewcommand{\onlynonthesis}[1]{#1}
\renewcommand{\onlythesis}[1]{}

\usepackage{float}
\usepackage{xcolor}
\usepackage{siunitx}
\usepackage{subcaption}

\newcommand{\sectionvspace}[1]{\vspace{#1}}

\def\etal{\emph{et al.\ }}

\renewcommand{\vec}{\mathbf}
\newcommand{\norm}[1]{\left\lVert #1 \right\rVert}
\newcommand{\normltwo}[1]{\norm{#1}_{2}^{2}}
\newcommand\figref{Fig.~\ref}
\newcommand\tabref{Table \ref}
\newcommand\equationref{Eq.~\ref}

\newcommand{\argmax}{\textit{argmax}}
\newcommand{\voxelspace}{V}  
\newcommand{\agentpose}{\vec{p}}  
\newcommand{\candidateposes}{P}
\newcommand{\occprior}{v^{o,\textit{prior}}}  
\newcommand{\Surf}{\textit{Surf}}  
\newcommand{\ObsSurf}{\textit{ObsSurf}}  
\newcommand{\score}{\textit{s}}  
\newcommand{\efficiency}{\textit{eff}}  
\newcommand{\obscountfactor}{\eta}  
\newcommand{\map}{M}  
\newcommand{\mapresolution}{r}  
\newcommand{\lasttime}{t_{e}}  

\usepackage[normalem]{ulem}

\newcolumntype{L}[1]{>{\raggedright\let\newline\\\arraybackslash\hspace{0pt}}m{#1}}
\newcolumntype{C}[1]{>{\centering\let\newline\\\arraybackslash\hspace{0pt}}m{#1}}
\newcolumntype{R}[1]{>{\raggedleft\let\newline\\\arraybackslash\hspace{0pt}}m{#1}}

\begin{document}

\title{Learn-to-Score: Efficient 3D Scene Exploration by Predicting View Utility}

\titlerunning{Learn-to-Score: Efficient 3D Scene Exploration by Predicting View Utility}

\authorrunning{B.~Hepp~et~al.}

\author{
	Benjamin Hepp\inst{1}\inst{2} \and
	Debadeepta Dey\inst{2} \and
	Sudipta N. Sinha\inst{2} \and \\
	Ashish Kapoor\inst{2} \and
	Neel Joshi\inst{2} \and
	Otmar Hilliges\inst{1}
}
\institute{
	ETH Zurich \and Microsoft Research
}

\maketitle

\begin{abstract}

Camera equipped drones are nowadays being used to explore large scenes and
reconstruct detailed 3D maps. When free space in the scene is approximately 
known, an offline planner can generate optimal plans to efficiently explore the 
scene. However, for exploring unknown scenes, the planner must predict and maximize
usefulness of where to go on the fly. Traditionally, this has been achieved using 
handcrafted utility functions. We propose to learn
a better utility function that predicts the usefulness of future viewpoints.
Our learned utility function is based on a 3D convolutional neural network.
This network takes as input a novel volumetric scene representation
that implicitly captures previously visited viewpoints and generalizes to new scenes.
We evaluate our method on several large 3D models of urban scenes using simulated depth cameras.
We show that our method outperforms existing utility measures in terms of reconstruction performance and is robust to sensor noise.

\end{abstract}




\section{Introduction}

Quadrotors, drones, and other robotic cameras are becoming increasingly powerful, inexpensive and are being  used for a range of tasks in computer vision and robotics applications such as autonomous navigation,  mapping, 3D reconstruction, reconnaissance, and grasping and manipulation. For these applications,
modeling the surrounding space and determining which areas are occupied 
is of key importance.

Recently, several approaches for robotic scanning of indoor \cite{xu2017autonomous} and outdoor \cite{roberts:2017,hepp2017plan3d} scenes have been proposed. Such approaches need to reason about whether voxels are free, occupied, or unknown space to ensure safety of the robot and to achieve good coverage w.r.t. their objective function (e.g. coverage of the 3D surfaces \cite{roberts:2017}).
Model-based approaches require approximate information about free space and occupied space, which is typically acquired
or input manually. This prevents such approaches from being fully autonomous or deployed in entirely unknown scenes
\cite{vasquez2014volumetric}.
Model-free approaches can be applied in unknown environments \cite{heng2015efficient,kriegel2015efficient,isler2016information,delmerico2017comparison}.
Irrespective of the type of approach used, all algorithms require a utility function that predicts how useful a new measurement (i.e.~depth image) would be.
Based on this utility function a planner reasons about the sequence of viewpoints to include in the motion plan.
This utility function is often a hand-crafted heuristic and hence it is difficult to
incorporate prior information about the expected distributions of 3D geometry in certain scenes.

We propose to devise a better utility function using a data-driven approach.
The desired target values for our utility function stem from an oracle with access to ground truth data.
Our learned utility function implicitly captures knowledge about building and geometry distributions
from approporiate training data and is capable of predicting the utility of new viewpoints
given only the current occupancy map.
To this end we train a 3D ConvNet on a novel multi-scale voxel representation of an underlying occupancy map, which encodes the current model of the environment.
We then demonstrate that the learned utility function can be utilized
to efficiently explore unknown environments.

The input to our network relies only on occupancy and hence abstracts away the capture method (i.e. stereo, IR depth camera, etc.).
While our training data consists of perfect simulated depth images
we demonstrate in our experiments that our learned model can be used with imperfect sensor data at test time, such as simulated noisy depth cameras or stereo data.
The approach is not limited
to environments with a fixed extent and generalizes to new scenes that are substantially different from ones in the training data.
Our approach outperforms existing methods, that use heuristic-based utility functions~\cite{vasquez2014volumetric,isler2016information,delmerico2017comparison}
and is more than $10 \times$ faster to compute than the methods from \cite{isler2016information,delmerico2017comparison}.


\section{Related work}
Exploration and mapping are well studied problems.
We first discuss theoretical results and then describe
approaches in the active vision domain
and finally work in 3D vision.

\vspace{-10pt}
\paragraph{Submodular sensor placement:}
In the case of a priori known environments and a given set of measurement locations, much work is dedicated to submodular objective functions for coverage
\cite{feige1998threshold,nemhauser1978analysis}.
Submodularity is a mathematical property
enabling approximation guarantees on the solution using greedy methods.
While work exists on dynamic environments where
the utility of future measurements can change upon performing
a measurement~\cite{golovin2011adaptive,hollinger2012active},
these methods are usually difficult to scale to large state and observation spaces,
which we considered in this
\onlynonthesis{paper}\onlythesis{chapter}
as they are common in computer vision applications.

\vspace{-10pt}
\paragraph{Next-best-view and exploration:}

In the next-best-view setting, the set of measurement locations is often fixed a priori as in the submodular coverage work described above.
The work in this area 
is usually concerned with defining good heuristic utility functions
and approximating the coverage task to make it computationally feasible~\cite{chen2011active,kriegel2015efficient,wenhardt2007active,forster2014appearance,dunn2009next}.
A number of heuristics is explicitly compared in~\cite{isler2016information,delmerico2017comparison},
and a subset of these is computed and used as a feature vector by Choudhury \etal~\cite{choudhury2017adaptive} to imitate an approximately optimal strategy with ground-truth access.

Based on an a priori fixed set of camera poses and a binary input mask
of already visited poses Devrim \etal~\cite{devrim2017reinforcement}
use reinforcement learning to regress a scalar parameter used in the
selection algorithm for the next view.
In contrast to our work the approach is concerned with
a priori known, fixed environments and camera poses
making it suitable for inspection planning.

In active vision, a large body of work is concerned with
exploration through only partially known scenes.
Frontier-based algorithms~\cite{yamauchi1997frontier} are used for autonomous mapping and exploration of the environment using stereo~\cite{fraundorfer2012vision}, RGB-D, or monocular cameras~\cite{shen2011autonomous}.
Heng \etal~\cite{heng2015efficient} propose a method which alternates
between exploration and optimizing coverage for 3D reconstruction.

All of the approaches discussed above either define or are based on heuristics to decide on the utility of the next measurement or require prior knowledge
of environment and possible camera poses.
Instead of hand-crafting a utility function our work is concerned
with learning such a function that can outperform existing hand-crafted functions
and is computationally cheaper to evaluate.
Additionally, our approach does not need a priori knowledge of the map.

\vspace{-10pt}
\paragraph{3D convolutional neural networks:}

A large body of work in computer vision is concerned with processing of
3D input and output data using convolutional neural networks.
In some cases this data stems from RGB-D images such as in
Song \etal~\cite{song2016deep} where the goal is to detect objects.
In other contexts, volumetric input in the form of binary
occupancy labels or signed distance functions are used
for diverse tasks such as shape classification
and semantic voxel labeling~\cite{dai2017scannet,Riegler2017CVPR},
surface completion~\cite{dai2016complete},
hand pose estimation~\cite{ge20173d},
or feature learning~\cite{zeng2017match}.
These works are concerned with passive tasks on uniform input grids of fixed
dimensions, containing the object or surface of interest.
This prevents reasoning across large distances or requires one to reduce
the level of detail of the input.

Different representations of occupancy grids have been proposed to mitigate the
trade-off of large uniform input dimensions and level of detail
\cite{Riegler2017CVPR}.
However, in the context of our work the occupancy map is often not very sparse
as it is generated by casting rays into a tri-state map and updating
continuous values which results in very few homogeneous regions which
would benefit from the formulation by Riegler \etal~\cite{Riegler2017CVPR}.
Also related to our work are approaches to multi-view reconstruction
\cite{choy20163d} where
the output is predicted based on a sequence of input images.
In contrast to our work Liu \etal~\cite{liu2015deep} reconstruct small objects in a fixed size volume
whereas we are concerned with large city scenes containing several buildings.


\section{Problem Setting and Overview}

\begin{figure}[ht]
	\begin{center}
		\includegraphics[width=1.0\linewidth]{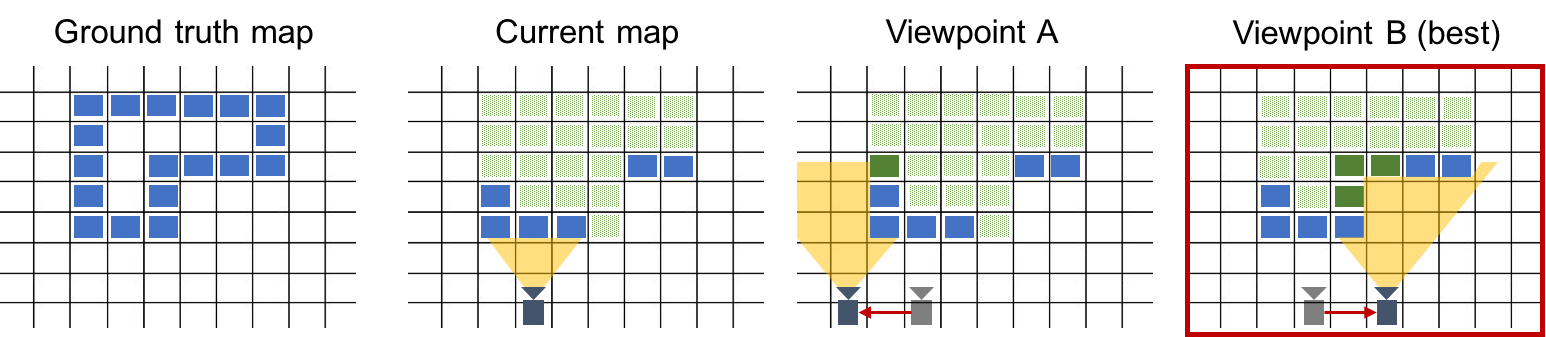}
	\end{center}
	\caption{
		The exploration task (here depicted in 2D for clarity)
		is to discover occupied surface voxels
		(shown here in blue). Voxels are
		initially unknown (shown here in light green) and get discovered by
		taking a measurement, e.g.,~shooting rays from the camera into the scene.
		Voxels that are not surface voxels will be discovered as free voxels
		(shown here in white).
		Each possible viewpoint has a corresponding utility value depending
		on how much it contributes to our knowledge of the surface
		(shown here in dark green).
		To decide which viewpoint we should go to next, an ideal utility score function
		would tell us the expected utility of viewpoints before performing them.
		This function can then be used in a planing algorithm to visit a sequence of viewpoints
		with the highest expected utility.
	}
	\label{fig:exploration_task_overview}
\end{figure}

Our work is concerned with the automatic exploration of an a priori unknown 3D world with the ultimate goal of reconstructing the surfaces of the scene in an efficient manner.
In this setting, illustrated in \figref{fig:exploration_task_overview}, an algorithm has to make decisions about the next viewpoint based only on the current map information.
In \figref{fig:exploration_task_overview} the camera is surrounded by some space, already known to be free (white) and part of the surface has been observed (blue). The next viewpoint is restricted to the known free space, whereas moving into unknown space (light green) could
lead to collisions.
The main difficulty stems from the fact that the algorithm needs to predict how much unknown surface can be discovered from a new viewpoint.
Much work has been dedicated to developing and studying  various heuristics to compute a score that quantifies the expected value of possible viewpoints~\cite{isler2016information,delmerico2017comparison}.

We propose a data-driven approach where we use supervised learning to find a utility function that imitates an oracle.
The oracle has access to the ground truth map and can compute the true utility score.
For this task we introduce a map representation consisting of multi-scale sub-volumes extracted around the camera's position.
For all possible viewpoints this data is fed into a 3D ConvNet at training time together with the oracle's score as a target value.
Intuitively, the model learns to predict the likelihood of seeing additional surface voxels for any given pose, given the current occupancy map.
However, we do not explicitly model this likelihood but instead provide only the oracle's score to the learner.
We experimentally show that our formulation 
generalizes well to new scenes with different object shape and distribution and can handle input resulting from noisy sensor measurements.

We follow related work~\cite{isler2016information,delmerico2017comparison,devrim2017reinforcement}
and evaluate our method on simulated but high-fidelity environments.
This allows for evaluation of the utility function and reduces the influence of  
environmental factors and specific robotic platforms.
Our environments contain realistic models of urban areas in terms of size and
distribution of buildings.
Furthermore it is important to note that our technique only takes occupancy information as input and does not directly interface with raw sensor data.
In addition we test our approach on real data from outdoor and indoor scenes to demonstrate that our method is not limited to synthetic environments.



\section{Predicting View Utility}
We first formally define our task and the desired utility function and then introduce our method for learning and evaluating this function.

\subsection{World model}
\label{sec:world_model}

We model the world as a uniform voxel grid $\voxelspace$ with resolution $\mapresolution$.
A map $\map$ is a tuple $\map = (\map^{o}, \map^{u})$ of functions
$\map^{o}: \voxelspace \rightarrow [0, 1]$, $\map^{u}: \voxelspace \rightarrow [0, 1]$
that map each voxel $v\in \voxelspace$ to an occupancy value $\map^{o}(v)$ describing the
fraction of the voxel's volume that is occupied and an associated uncertainty value $\map^{u}(v)$,
i.e.~$1$ for total uncertainty and $0$ for no uncertainty.
Maps change over time so we denote the map at time $t$ as $\map_{t}$.

After moving to a viewpoint $\agentpose$ the camera acquires a new measurement
in the form of a depth image
and the map $\map$ is updated. We denote the updated map as $\map|_{\agentpose}$.
The uncertainty is updated according to
\begin{align}
\map^{u}|_{\agentpose}(v) = \exp(-\obscountfactor) \map^{u}(v) \quad ,
\label{eq:uncertainty_update}
\end{align}
where $\obscountfactor \in \mathbb{R}_{>0}$ describes the
amount of information added by a single measurement.
This is a simple but effective measure providing a diminishing information
gain of repeated measurements.
Note that $\map^{u}|_{\agentpose}(v) \leq \map^{u}(v)$ so uncertainty decreases with
additional measurements.
As is typical in occupancy
mapping \cite{thrun2005probabilistic,hornung13auro}
we update the voxel occupancies $\map^{o}(v)$
according to a beam-based inverse sensor model.
Please see Sec.~\ref{sec:3d_scene_exploration} for details on initialization
of the map.

\subsection{Oracle utility function}
To select viewpoints, we need a utility function that assigns scores to all possible viewpoints at any time.
We first introduce an oracle utility function with access to the ground truth
(set of true surface voxels) during evaluation. It returns the desired true utility measure.
We will then learn to imitate the oracle without access
to ground truth.

We characterize a good viewpoint  as one that discovers a large amount of
surface voxels.
Let $\ObsSurf(\map)$ be the total number of observed surface voxels in map $\map$ weighted by their associated certainty value:
\begin{align}
\ObsSurf(\map) = \sum_{v \in \textit{Surf}} (1 - \map^{u}(v)) \quad ,
\end{align}
where $\Surf \subseteq V$ is the set of ground truth surface voxels,
i.e.~all voxels that intersect the surface.
Note that $\ObsSurf(\map)$ increases monotonically with additional measurements
because the certainty of voxels can only increase according to Eq. \eqref{eq:uncertainty_update}.

The decrease in uncertainty of surface voxels with a new measurement defines the oracle's utility score.
We express this score as a function of the current map $\map$
and the camera pose $\agentpose$:
\begin{align}
\score(\map, \agentpose) & = \ObsSurf(\map|_{\agentpose}) - \ObsSurf(\map) \nonumber \\
& = \sum_{v \in \Surf} \left( - \map^{u}|_{\agentpose}(v) + \map^{u}(v) \right)
= \sum_{v \in \Surf}  \left( 1 - \exp(-\obscountfactor) \right) \map^{u}(v) \geq 0 \quad .
\label{eq:oracle-utility}
\end{align}

\subsection{Learning the utility function}

Computing the utility score introduced in \equationref{eq:oracle-utility} for any viewpoint
requires access to the ground truth map.
Our goal is to predict $\score(\map, \agentpose)$ without access to this data
so we can formulate a regression problem that computes score values given occupancy maps as input.

\subsubsection{Multi-scale map representation}
We propose to make predictions directly based on the occupancy map, rather than based on a temporal sequence of raw inputs.
This occupancy map encodes our knowledge of already observed surfaces and free space and ultimately can be used to build
up a map for both navigation and 3D reconstruction.

For use in a 3D ConvNet the map has to be represented with fixed dimensionalities. Here a trade-off between memory consumption, computational cost, reach and resolution arises. For example, extracting a small high resolution grid around the camera would constrain information to a small spatial extent whereas a grid with large spatial extent would either lead to rapid increase in memory consumption and computational cost or would lead to drastic reduction in resolution.

\begin{figure}[h]
	\begin{center}
		\includegraphics[width=0.7\linewidth]{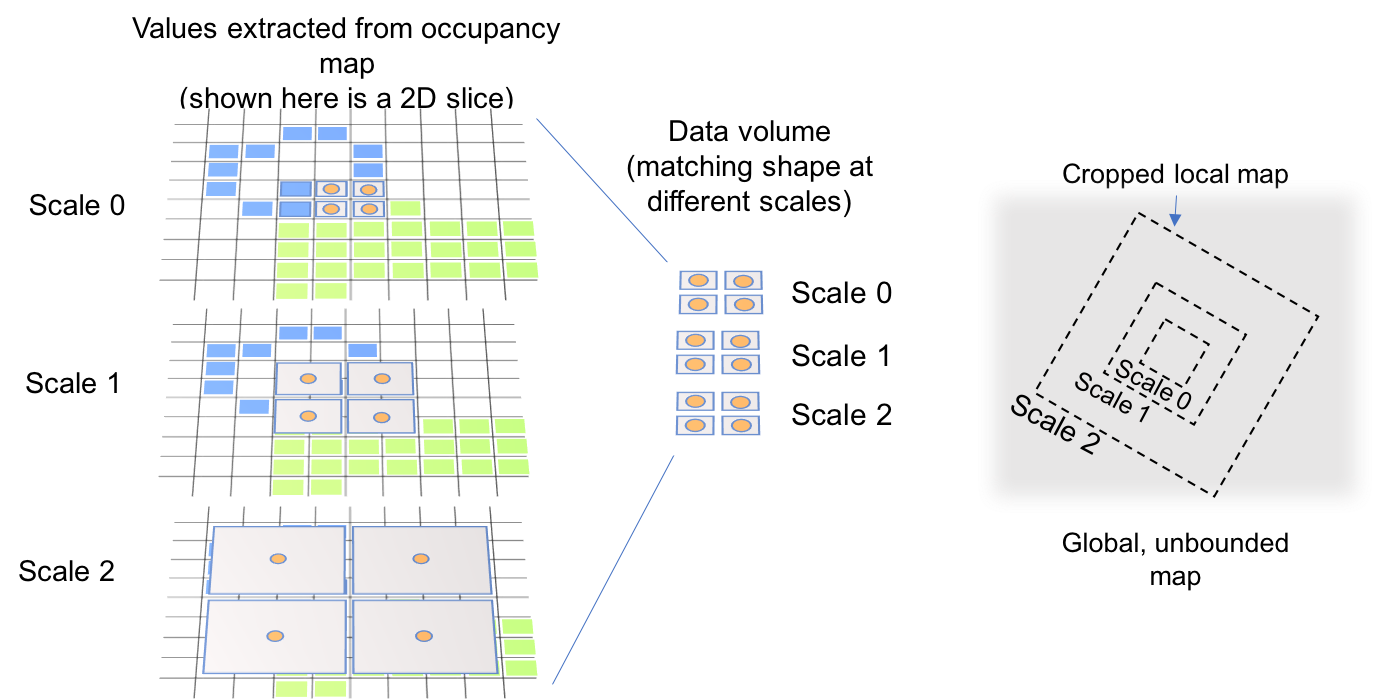}
	\end{center}
	\caption{Local multi-scale representation of an occupancy map.
		For clarity of presentation we shows the 2D case for a
		grid of size $2 \times 2$.
		The occupancy map is sampled with 3D grids at multiple scales centered around the camera position.
		Sample points on different scales are shown in orange
		and their extent in gray.
		\vspace{-10pt}
	}
	\label{fig:occupancy_grid_extraction}
\end{figure}

To mitigate this issue we introduce a multi-scale representation
by sampling the occupancy map at multiple scales as depicted in \figref{fig:occupancy_grid_extraction}.
For each scale $l \in \{1, \ldots, L\}$ we extract values on a 3D grid of size
$D_{x}\times D_{y}\times D_{z}$ and resolution $2^{l} \mapresolution$
(orange points in \figref{fig:occupancy_grid_extraction}).
On scale $l$ the map values are given by averaging the $2^{l}$ closest voxels
(gray rectangles in \figref{fig:occupancy_grid_extraction}).
This can be done efficiently by representing the map as an octree.
The 3D grids are translated and rotated
according to the measurement pose $\agentpose$ and we use tri-linear interpolation of the map values to compute the values on the grid. This representation allows us to capture both coarse parts of the
map that are far away from the camera but still keep finer detail in its direct surroundings. Furthermore, it provides an efficient data representation of fixed size, suitable for training of a 3D ConvNet.
We denote the multi-scale representation by
$x(\map, \agentpose) \in \mathbb{R}^{D_{x}\times D_{y}\times D_{z}\times 2 L}$.
Note that the factor $2$ stems from extracting the occupancy and the uncertainty value on each scale.

\subsubsection{ConvNet Architecture}
\label{sec:convnet_architecture}

\begin{figure}[h]
	\begin{center}
		\includegraphics[width=0.6\linewidth]{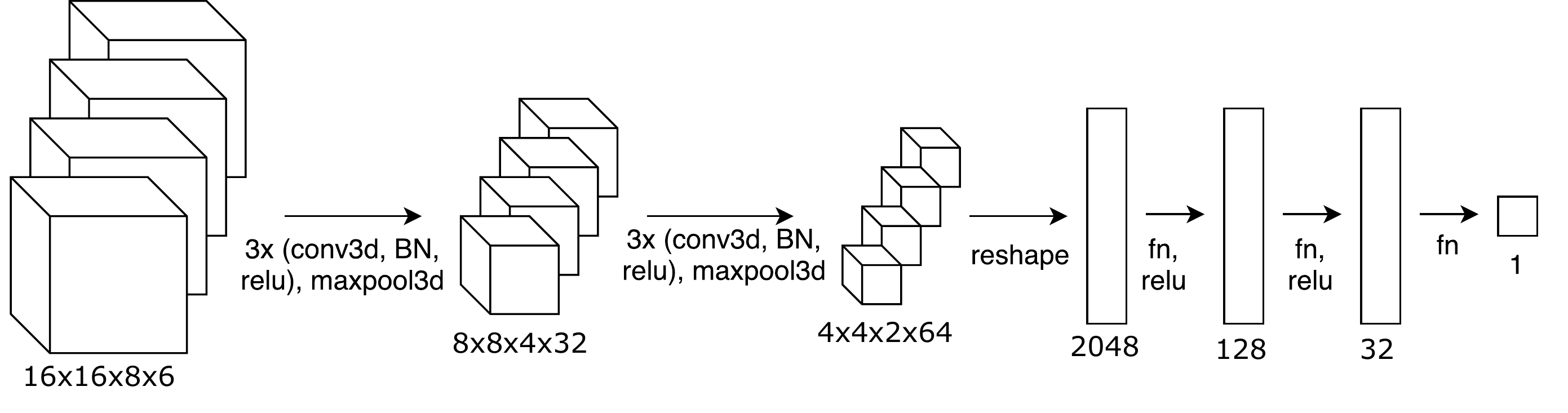}
	\end{center}
	\caption{
		Our architecture for an input size
		of $16\times 16\times 8$ with $L=3$ scales resulting in
		$2 L = 6$ channels.
		The model
		consists of blocks (made up of multiple units
		each performing 3D convolution, batch-norm and ReLu)
		followed by downscaling using 3D max-pooling. This pattern is performed
		until we arrive at a data volume with spatial dimension
		$4\times 4\times 2$. This is reshaped
		into a single vector followed by two fully connected layers with ReLu activation
		and a final linear layer predicting a scalar score value.
	}
	\label{fig:conv_net}
\end{figure}

We now describe our proposed model architecture used to learn the desired utility function
$f: \mathbb{R}^{D_{x}\times D_{y}\times D_{z}\times 2 L} \rightarrow \mathbb{R}$.
The general architecture is shown in~\figref{fig:conv_net} and
consists of a number $N_{c}$ of convolutional blocks
followed by two fully connected layers with ReLu activations.
Each convolutional block contains a series of $N_{u}$ units
where a unit is made up of a 3D convolution, followed by Batch-Norm, followed
by ReLu activation. Each 3D convolution increases the number of feature
maps by $N_{f}$. After each block the spatial dimensions are downscaled by a
factor of $2$ using 3D max-pooling.
The first fully connected layer has $N_{h1}$ hidden units and
the second one has $N_{h2}$ hidden units.
Note that we do not separate the input data at different scales so that
the network can learn to combine data on different scales.
More details on the exact architecture are provided in
Sec.~\ref{sec:convnet_experiments} and an evaluation of different variants
is provided in Supplementary Material.

We use a weight-regularized $L2$ loss
\begin{align}
\mathcal{L}(X, Y; \theta) = \sum_{i=1}^{N}
	\normltwo{f(X_{i}) - Y_{i}} + \lambda \normltwo{\theta}
	\quad ,
\label{eq:loss}
\end{align}
where
$\theta$ are the model parameters,
$\lambda$ is the regularization factor
and $(X_{i}, Y_{i})$ for $i \in \{1, \ldots, N\}$
are the samples of input and target from our dataset. 

\subsection{3D Scene Exploration}
\label{sec:3d_scene_exploration}

To evaluate the effectiveness of our utility function, we implement a next-best-view (NBV) planning approach, to 
sequentially explore a 3D scene.
Here we provide details of our world model and
our methods for execution of episodes for the data generation phase and at test time.

We assume exploration of the world occurs in episodes.
To initialize a new episode, the camera pose at time $t_{0}$ is chosen randomly in free space
such that no collision occurs and the camera can move to each neighboring viewpoint without collision.
A collision occurs if a bounding box of size $(1m, 1m, 1m)$
centered at the camera pose
intersects with any occupied or unknown voxel.
Initially, all voxels $v \in \voxelspace$ are initialized to be unknown,
i.e.~$\map_{t_{0}}^{u}(v) = 1, \map_{t_{0}}^{o}(v) = \occprior \  \forall v \in \voxelspace$,
where $\occprior$ is a prior assumption on the occupancy and we use
$\occprior = 0.5$ throughout this work.
To enable initial movement of the camera we clear (i.e.~set to free space)
a bounding box of $(6 m)^{3}$ around the initial camera position.

At each time step $t$, we evaluate each potential viewpoint with our utility function, and
move to the viewpoint
$\agentpose^{*}(t)$ that gives the best expected reward according to:
\begin{align}
\agentpose^{*}(t) = \argmax_{\agentpose \in \candidateposes(t)} u(\map_{t}, \agentpose) \quad,
\label{eq:greedy_action_selection}
\end{align}
where $\candidateposes(t)$ is the set of potential viewpoints
and $u(\cdot)$ is the utility function in use.

At the start of each episode the set of potential viewpoints only contains the initial viewpoint.
At each time step the set $\candidateposes(t)$
is extended by those neighbors of the current viewpoint
that do not lead to a collision.
We ignore potential viewpoints if they have already been observed twice.
Each viewpoint has $9$ neighbors,
$6$ of them being positive and negative translations of $2.5m$ along
each axis of the camera frame, $2$ rotations of $\ang{25}$, clock-wise
and counter-clockwise along the yaw axis, and a full turnaround
of $\ang{180}$ along the yaw axis. We keep pitch and roll angles
fixed throughout.

After moving to a new viewpoint, the camera takes a measurement in the form
of a depth image and the map is updated
(see Supplementary Material for details on the camera parameters and the map update).
Note that we use ground truth depth when generating training data
but later demonstrate that we can use noisy depth images and even stereo depth
at test time.

Note that we assume that the utility function is submodular. While this is true for the oracle utility
it is not necessarily the case for other utility functions (i.e.~our learned model). Nevertheless, this assumption
allows us to perform lazy evaluations of the utility function \cite{krause2012submodular} (see Supplementary Material for details).

\subsection{Dataset}
\label{sec:dataset}

\begin{figure*}
	\centering
	\includegraphics[width=0.16\linewidth]{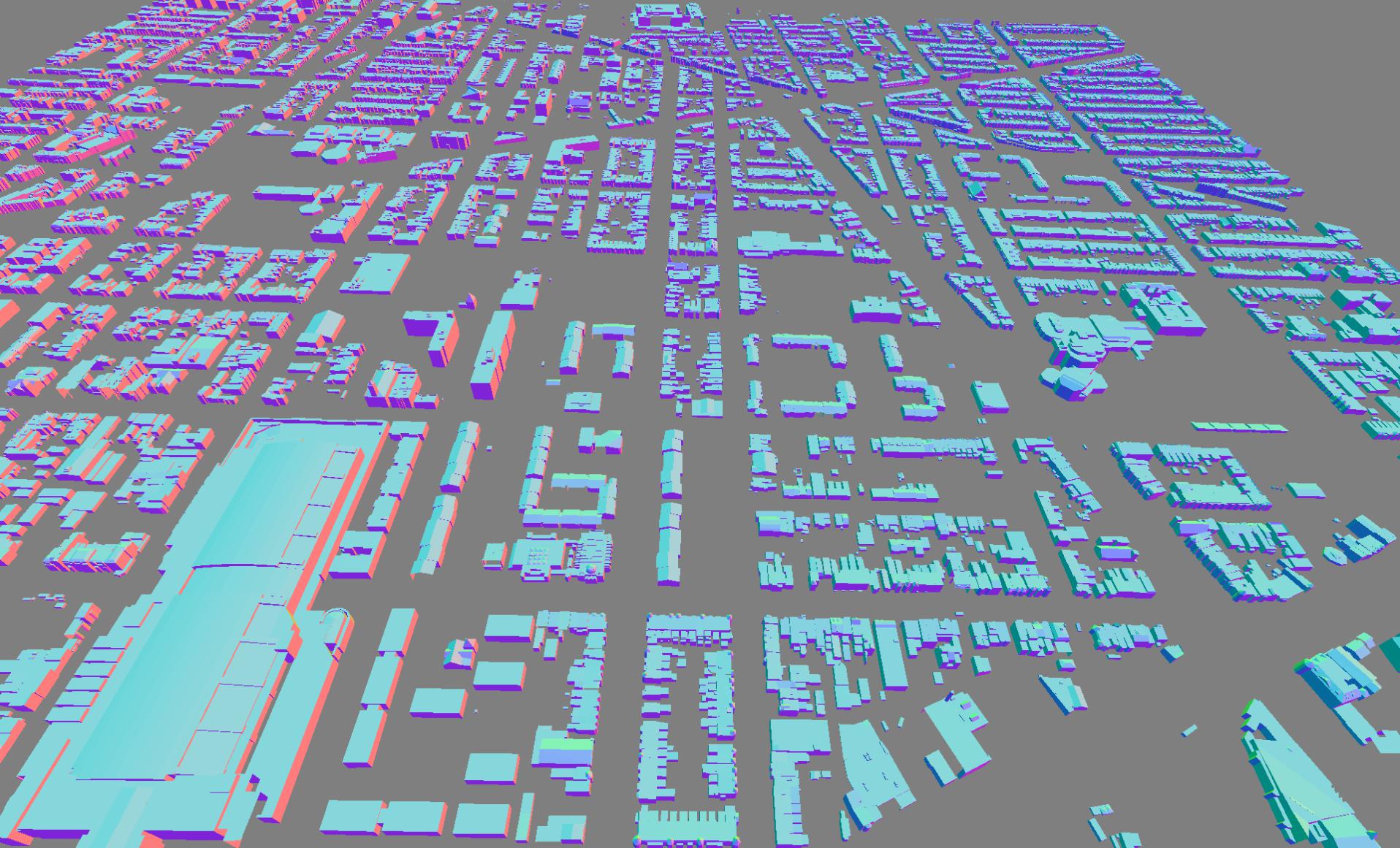}
	\includegraphics[width=0.16\linewidth]{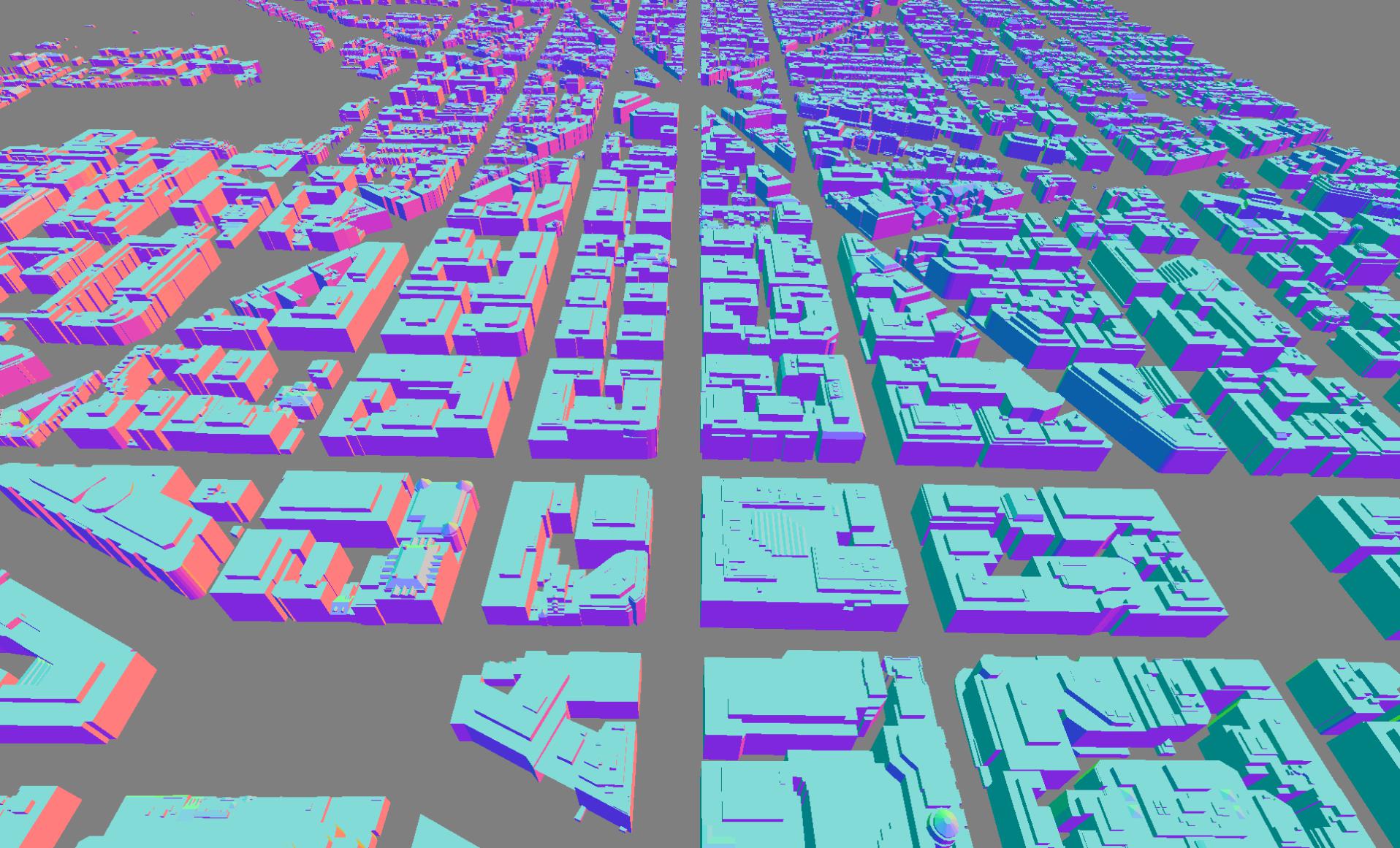}
	\includegraphics[width=0.16\linewidth]{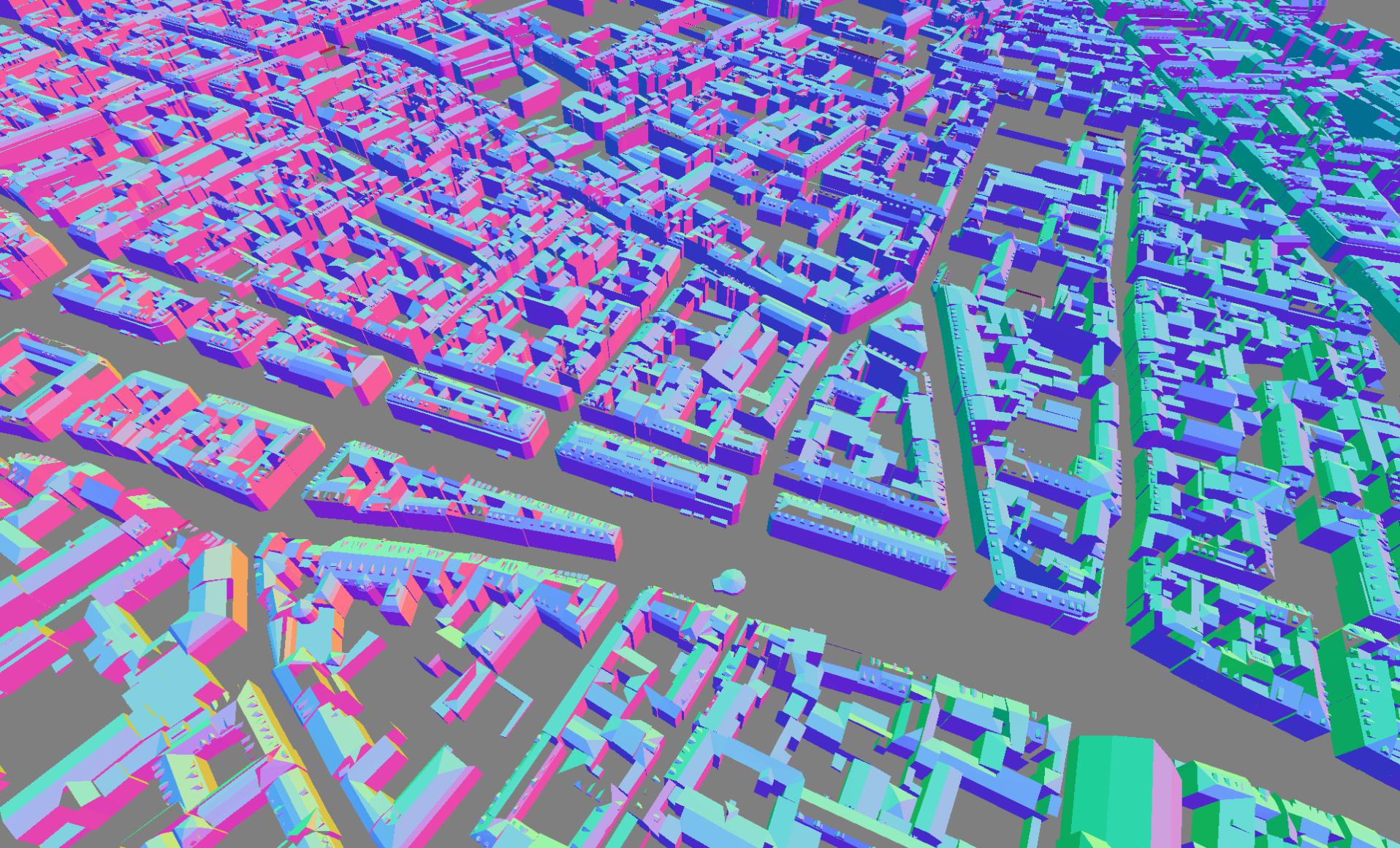}
	\includegraphics[width=0.16\linewidth]{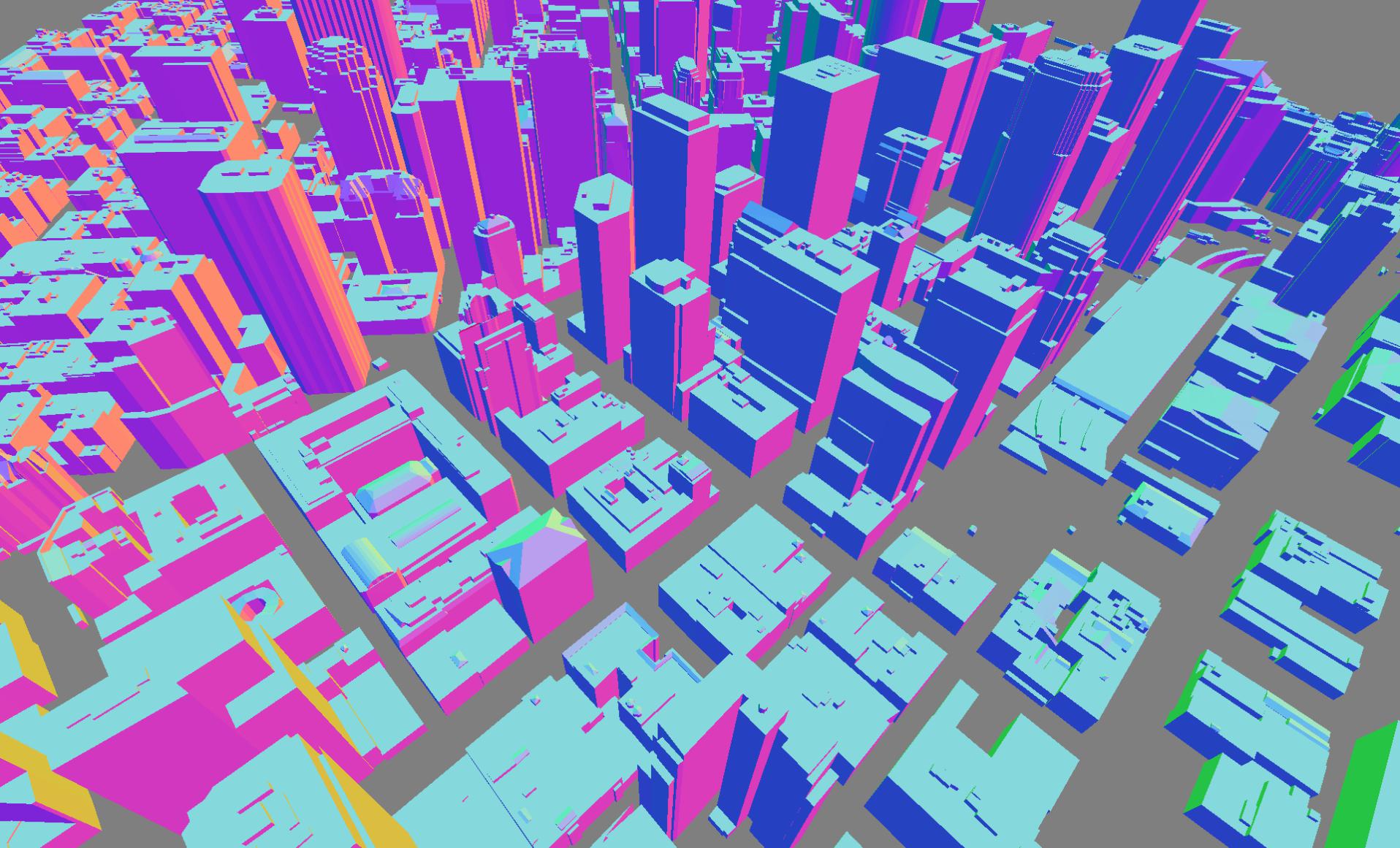}
	\includegraphics[width=0.16\linewidth]{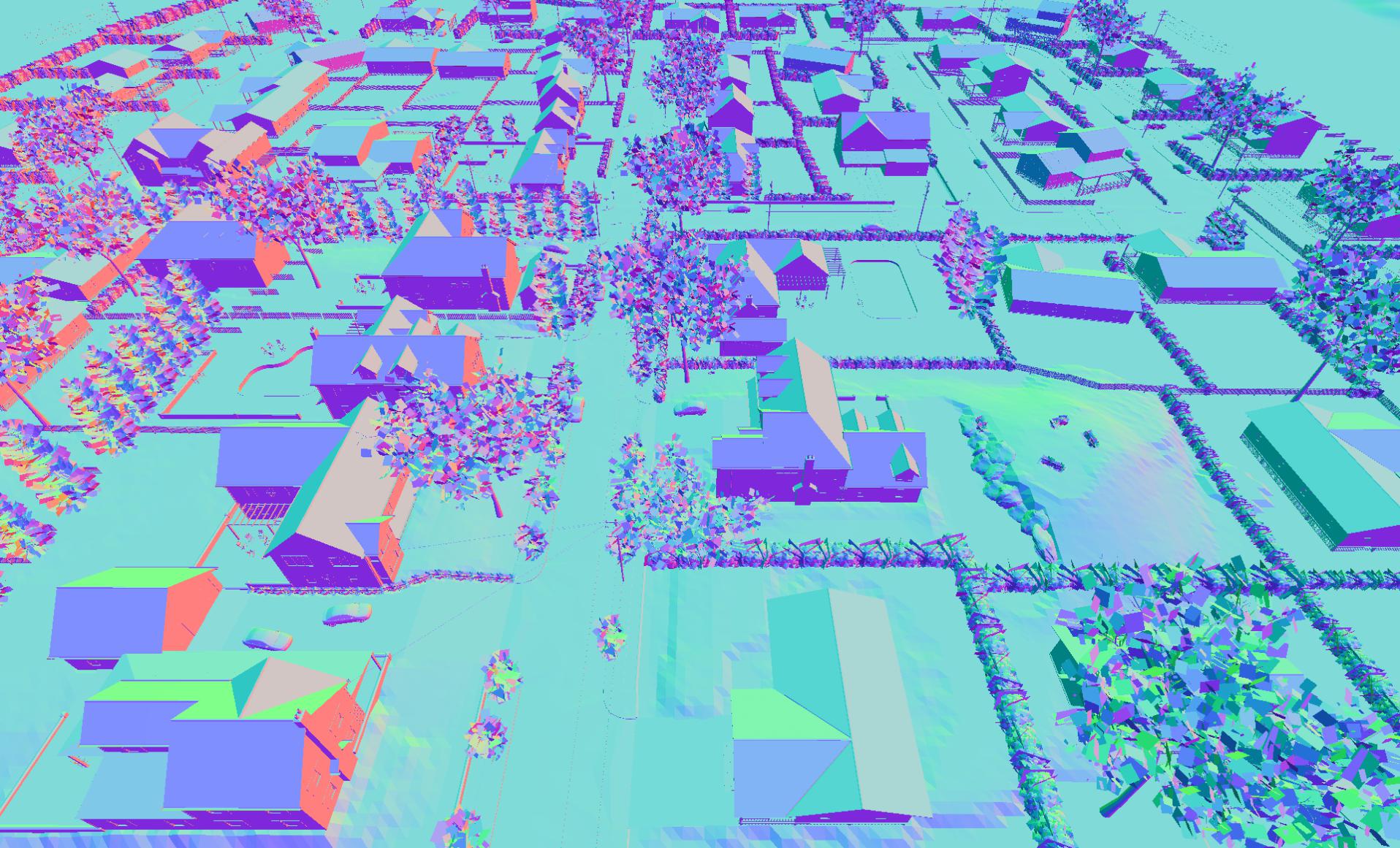}
	\caption{
		Normal rendering of environments.
		From left to right: \textit{Washington2}, \textit{Washington1},
		\textit{Paris}, \textit{SanFrancisco}, \textit{Neighborhood}. \vspace{-5pt}
	}
	\label{fig:scenes}
\end{figure*}

To learn the utility function
$f(x)$, approximating the oracle (see \equationref{eq:oracle-utility}) we require labeled training data.
Our data should capture large urban environments with a variety of structures
typical for human-made environments.
To this end we chose models from the \textit{3D Street View} dataset~\cite{zamir2016generic}.
These models feature realistic building distributions
and geometries from different cities. Additionally, we chose a large
scene from a photo-realistic game engine (\url{https://www.unrealengine.com})
containing small buildings in a suburban
environment, including trees, smaller vegetation and power lines.
All environments are shown in \figref{fig:scenes}.
Note that we only use data from \textit{Washington2} to train our
model. While \textit{Washington1} and \textit{Paris} are similar in terms of building height
the building distribution and geometry is clearly different.
A particular challenge is posed by the \textit{SanFrancisco} scene which includes
tall buildings never seen before in \textit{Washington2}.
Similarly, the buildings and vegetation in the \textit{Neighborhood} scene are
unlike anything seen in the training data.

We generate samples by running episodes with
$\mapresolution = 0.4m$ until time $\lasttime = 200$
and selecting the best viewpoint $\agentpose$ according to the oracle's score at each time step.
For each step $t$ we store tuples of input $x(\map, \agentpose)$ and target value from the oracle $s(\map, \agentpose)$
for each neighboring viewpoint.

Note that we record samples for each possible neighbor of the current viewpoint
(instead of only the best selected viewpoint).
This is necessary as our predictor will have to provide
approximate scores for arbitrary viewpoints at test time.
We record a total of approximately $1,000,000$ samples and
perform a $80/20$ split into training and validation set.
To encourage future comparison we will release our code for generating the data
and evaluation.
\sectionvspace{-10pt}


\section{Experiments}

We describe our ConvNet architecture and then show different evaluations of our method.

\subsection{ConvNet architectures and training}
\label{sec:convnet_experiments}

We evaluated different ConvNet variants by varying
$N_{c}$, $N_{u}$ and $N_{f}$.
We also tried modifications such as using residual units
\cite{He_2016_CVPR,he2016identity}. We report these results
in the supplementary material. Here we report results on the
best performing model with input size
$16\times 16\times 8$ ($N_{c} = 2$, $N_{u} = 4$, $N_{f} = 8$, $L=3$,
$N_{h1}=128$, $N_{h2}=32$),
denoted as \textbf{Ours} in the rest of the section.
Training of the model is done with ADAM using a mini-batch size of $128$,
regularization $\lambda = 10^{-4}$, $\alpha = 10^{-4}$
and the values suggested by Kingma \etal~\cite{kingma2014adam} for the other parameters.
Dropout with a rate of $0.5$ is used after fully-connected layers during training.
Network parameters are initialized according to Glorot \etal~\cite{glorot2010understanding} (corrected for ReLu activations).
We use early stopping when over-fitting on test data is observed.

\subsection{Evaluation}

Our evaluation consists of three parts.
First we evaluate our model on datasets generated as described in Sec.~\ref{sec:dataset}
and report spearman's rho to show the rank correlation of predicted scores and ground truth scores.
Following this, we compare our
models with previously proposed utility functions from~\cite{vasquez2014volumetric,isler2016information,delmerico2017comparison}.
We use the open-source implementation provided by~\cite{isler2016information,delmerico2017comparison}
and report results on their best performing methods on our scenes, \textit{ProximityCount} and \textit{AverageEntropy}.
We also compare with a frontier-based function measuring the number of unobserved voxels visible from a viewpoint
as in~\cite{heng2015efficient,bircher2016receding}.
For this comparison we use simulated noise-free depth images for all methods.
Finally, we evaluate our models with depth images perturbed by noise
and depth images produced by stereo matching in a photo-realistic
rendering engine.

To demonstrate the generalization capability of our models
we use four test scenes (column 2-5 in \figref{fig:scenes})
that show different building distribution and geometry
than the scene used to collect training data.
We also perform the experiments on the training scenes
where the exploration remains difficult due to random start poses
and possible ambiguity in the incomplete occupancy maps.

To compute score and efficiency values, we run $50$ episodes
with $\mapresolution = 0.4m$ until $\lasttime = 200$
for each method and compute the sample mean and standard deviation
at each time step. To enable a fair comparison, we select a random
start position for each episode in advance and use the same start positions
for each method.

In order to report a single metric of performance for comparison we compute the area under the curve of observed surface versus time (see plots in \figref{fig:example_scenes_and_maps}):
\begin{align}
\efficiency = \sum_{t=0}^{\lasttime} \ObsSurf(\map_{t}) \quad .
\label{eq:efficiency}
\end{align}
We call this metric \textit{Efficiency} as it
gives a higher score to a method that discovers surface early on.

\subsection{Model performance on different datasets}

Here we evaluate the performance of our model on data collected from different scenes
as described in Sec. \ref{sec:dataset}.
The model was trained on the training set of \textit{Washington2} and we report
Spearman's rho as well as the loss value from Eq. \eqref{eq:loss}
in Tab. \ref{tab:spearman_loss_results}.

\begin{table*}
	\centering
	\begin{tabular}{rcccccc}
		\hline
		\multicolumn{7}{c}{Evaluation on different datasets} \\
		& \begin{tabular}{@{}c@{}}\emph{Washington2}\\train\end{tabular}
		& \begin{tabular}{@{}c@{}}\emph{Washington2}\\test\end{tabular}
		& \emph{Washington1}
		& \emph{Paris}
		& \emph{SanFrancisco}
		& \emph{Neighborhood} \\
		\hline
		Spearman's rho
		& 0.88
		& 0.87
		& 0.83
		& 0.69
		& 0.73
		& 0.48 \\
		Loss value
		& 0.25
		& 0.28
		& 0.43
		& 0.63
		& 0.60
		& 0.93 \\
		\hline
	\end{tabular}
	\caption{
		Spearman's rho and loss values for our model on the different datasets.
		Despite the different building distribution and geometry of the test scenes
		(i.e.~all scenes but \textit{Washington2})
		compared to training data Spearman's rho value shows a high
		rank correlation with the oracle score. This is even the case
		for the \textit{Neighborhood} scene which features building shapes and trees
		unlike any in the training data.
	}
	\label{tab:spearman_loss_results}
\end{table*}

The Spearman's rho shows a clear rank correlation even for the \textit{Neighborhood}
scene which features building distribution and geometry significantly different from \textit{Washington2} which was
used to generate training data.
Interestingly, the model shows a high rank correlation for the \textit{SanFrancisco} scene which features tall
buildings and thus requires our model to generalize to different occupancy map distributions at high viewpoints.

\subsection{Comparison with baselines}

\begin{figure*}[!ht]
	\begin{center}
		\includegraphics[width=0.8\linewidth]{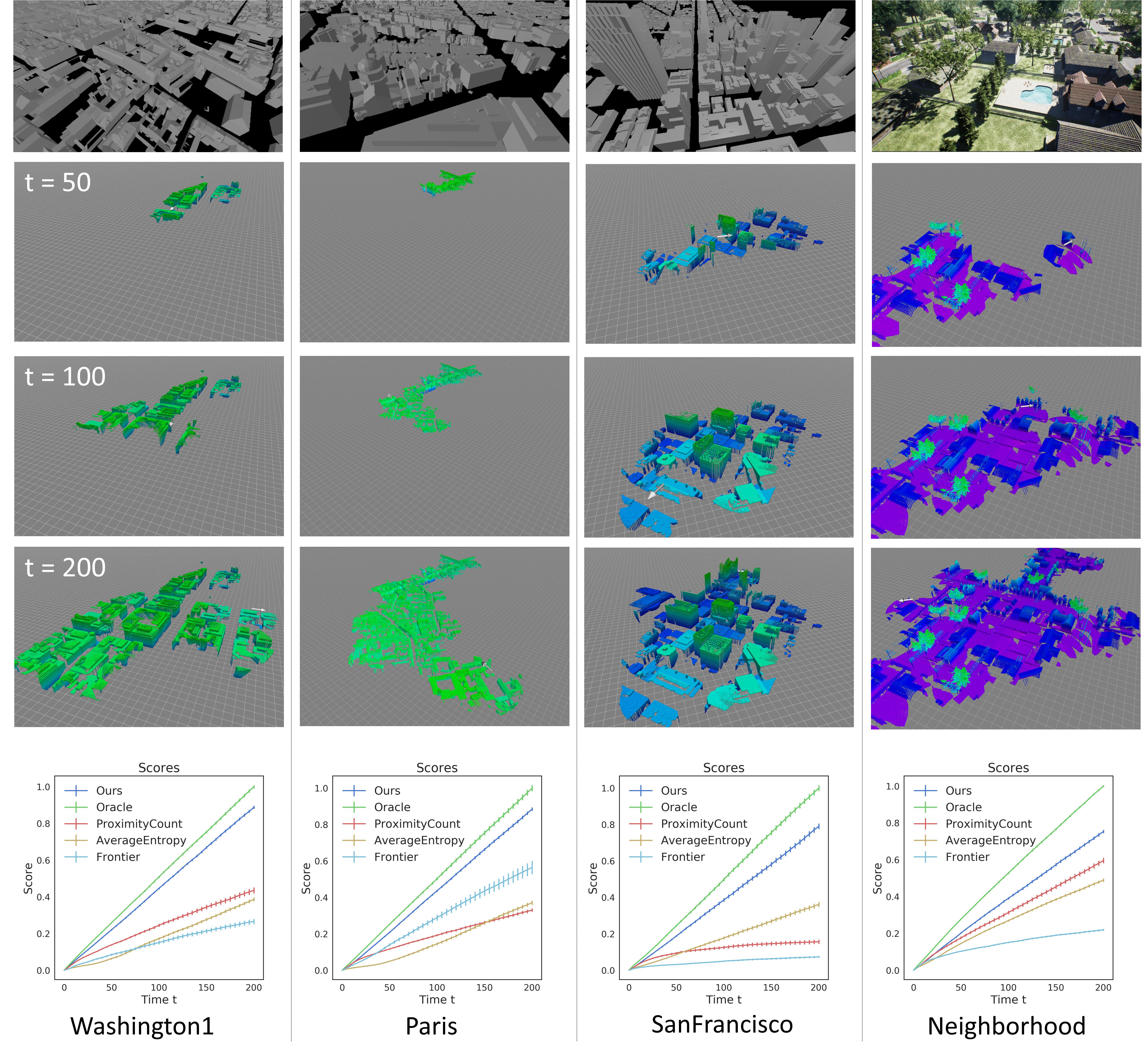}
	\end{center}
	\caption{
		Results on all test scenes.
		Top row: Visualization of the underlying mesh model.
		Row 2-4: Reconstructed 3D models at different time steps. Shown are only
		occupied voxels and the color coding indicates the voxel position along the z-axis.
		Bottom row: Plot of observed surface voxels vs.~time for all methods,
		the oracle with access to ground truth and the baseline methods.
		Our method performs the best and approaches the oracle's performance.
		Best viewed in color and zoomed in. Larger versions in Supplementary Material.
	}
	\label{fig:example_scenes_and_maps}
\end{figure*}

In \tabref{tab:main_model_vs_baselines} we compare the performance of our models against related hand-crafted utility functions~\cite{vasquez2014volumetric,isler2016information,delmerico2017comparison}.
Our method consistently
outperforms the existing functions in terms of the efficiency measure,
and as shown in \tabref{tab:computation_times_comparison},
is faster to compute than other methods.

\begin{table*}
	\centering
	\begin{tabular}{lccccc}
		\hline
		\multicolumn{6}{c}{Evaluation on different scenes} \\
		& \emph{Washington2} & \emph{Washington1}
		& \emph{Paris} & \emph{SanFrancisco}
		& \emph{Neighborhood} \\
		\hline

		Frontier
		& $0.40$ & $0.29$ & $0.57$ & $0.09$ & $0.27$ \\
		AverageEntropy~\cite{isler2016information}
		& $0.26$ & $0.36$ & $0.32$ & $0.30$ & $0.50$ \\
		ProximityCount~\cite{isler2016information}
		& $0.52$ & $0.47$ & $0.37$ & $0.23$ & $0.60$ \\
		\textbf{Ours}
		& $\mathbf{0.91}$ & $\mathbf{0.88}$ & $\mathbf{0.87}$
		& $\mathbf{0.77}$ & $\mathbf{0.74}$ \\
		{\color{gray}Oracle (GT access)}
		& \color{gray}$1.00$ & \color{gray}$1.00$ & \color{gray}$1.00$ & \color{gray}$1.00$ & \color{gray}$1.00$ \\
		\hline
	\end{tabular}
	\caption{
		Comparison of \textit{Efficiency} metric. Our method achieves a higher value
		than the other utility functions on all scenes
		showing that our learned models can generalize to other scenes.
		Note that the model is trained only on data recorded from \textit{Washington2}.
		\textit{Efficiency} values are normalized with respect to the oracle for easier comparison.
	}
	\label{tab:main_model_vs_baselines}
\end{table*}

We also show plots of observed surface voxels vs.~time for our model,
the oracle with access to ground truth
and baseline methods in \figref{fig:example_scenes_and_maps}.
Note that the scenes shown have not been used to generate any training data.
The results show that our method performs better compared to the baseline methods
at all time steps. Additionally the behavior of our method is consistent over all scenes
while the performance of the baselines varies from scene to scene.
The progression of reconstructed 3D models is shown by the renderings of the occupancy map at different times.

\begin{table*}
	\centering
	\begin{tabular}{lcccc}
		\hline
		\multicolumn{5}{c}{Computation time per step} \\
		& \ Frontier\  & \ ProximityCount\  & \ AverageEntropy\  & \ Ours\  \\
		\hline
		Time in s & 0.61 & 5.89 & 8.35 & \textbf{0.57}\\
		\hline
		
	\end{tabular}
	\caption{
		Comparison of computation time per step.
		Our method is as fast as a simple raycast in the \textit{Frontier} method
		and more than $10 \times$
		faster than \textit{ProximityCount} and \textit{AverageEntropy}.
	}
	\label{tab:computation_times_comparison}
\end{table*}

\subsection{Noisy input sensor}

While all our training is done on simulated data using ground truth depth images
our intermediate state representation as an occupancy map makes our models
robust to the noise characteristics of the input sensor.
Here we evaluate the performance of our models at test time with
depth images perturbed by noise of different magnitude.
Additionally we test our models with depth images computed from a
virtual stereo camera. To this end
we utilize a photorealistic game engine to synthesize RGB stereo pairs
and compute depth maps with semi-global matching.

Episodes were run with noisy depth images and the viewpoint sequence
was recorded. We replayed the same viewpoint sequences and used ground truth depth images
to build up an occupancy map and measure the efficiency.
Resulting \textit{Efficiency} values are listed in \tabref{tab:depth_noise_evaluation}.
One can see that our method is robust to different noise levels.
More importantly, even with depth images from the virtual stereo camera,
resulting in realistic perturbations of the depth images (see Supplementary Material),
our method does not degrade.

\begin{table*}
	\centering
	\begin{tabular}{lcccccc}
		\hline
		\multicolumn{7}{c}{Evaluation using noisy depth images (normalized)} \\
		\hline
		Noise & \ none\  & \ low\  & \ medium\  & \ high\  & \ very high\  & \ stereo\  \\
		$\efficiency$
		
		& $1.00$
		& $0.99$
		& $1.01$
		& $0.99$
		& $1.02$
		& $0.99$
		\\
		\hline
	\end{tabular}
	\caption{
		Comparison of our method using noisy depth images.
		\textit{Efficiency} values are normalized to the noise-free case.
		For the noise cases $40\%$ of pixels in each depth image were dropped
		and each remaining pixel was perturbed by normal noise
		($\sigma = 0.1m$ for low, $\sigma = 0.2m$ for medium, $\sigma = 0.5m$
		for high, $\sigma = 1.0m$ for very high).
		In the case of stereo matching we used a photo realistic rendering engine
		to generate stereo images with a baseline of $0.5m$. A disparity
		and depth image was computed using semi global matching~\cite{hirschmuller2008stereo}.
		Note that all values have a standard deviation of $\approxeq 0.03$.
	}
	\label{tab:depth_noise_evaluation}
\end{table*}

\subsection{Additional results on real data}

\begin{figure}[h]
	\centering
	\includegraphics[width=0.75 \linewidth]{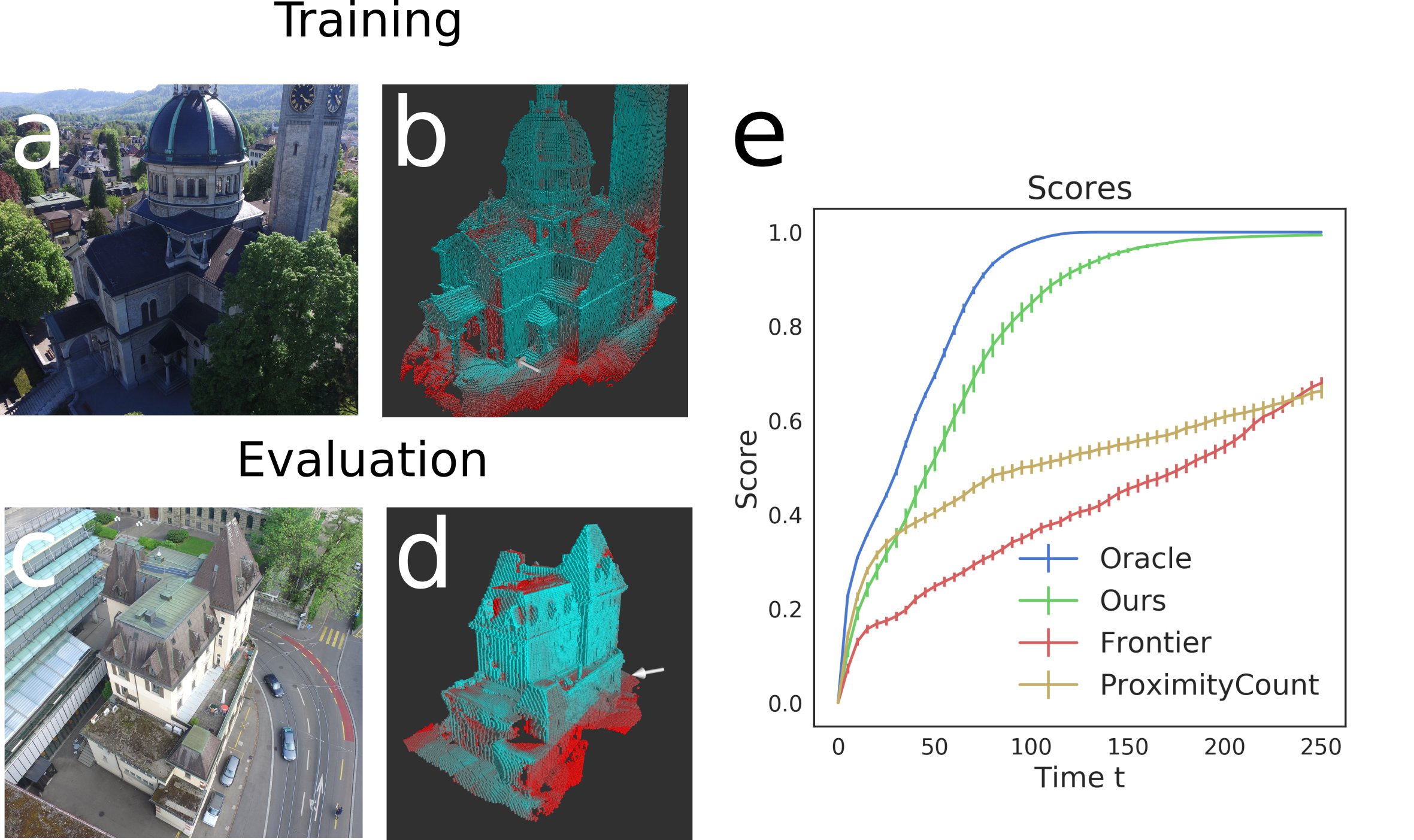}
	\caption{
		Shown are example explorations on \emph{real} outdoor data --
		(a) Picture of church scene.
		(b) Occupancy map of the church scene (training data) (200 steps).
		(c) Picture of historic building scene.
		(d) Occupancy map of the historic building scene (evaluation) (100 steps).
		(e) Performance plot for the historic building scene.
		Color coding of observed voxels: High uncertainty (red) and low uncertainty (cyan).
	}
	\label{fig:additional_results_outdoor}
\end{figure}

\begin{figure}[h]
	\centering
	\includegraphics[width=0.7 \linewidth]{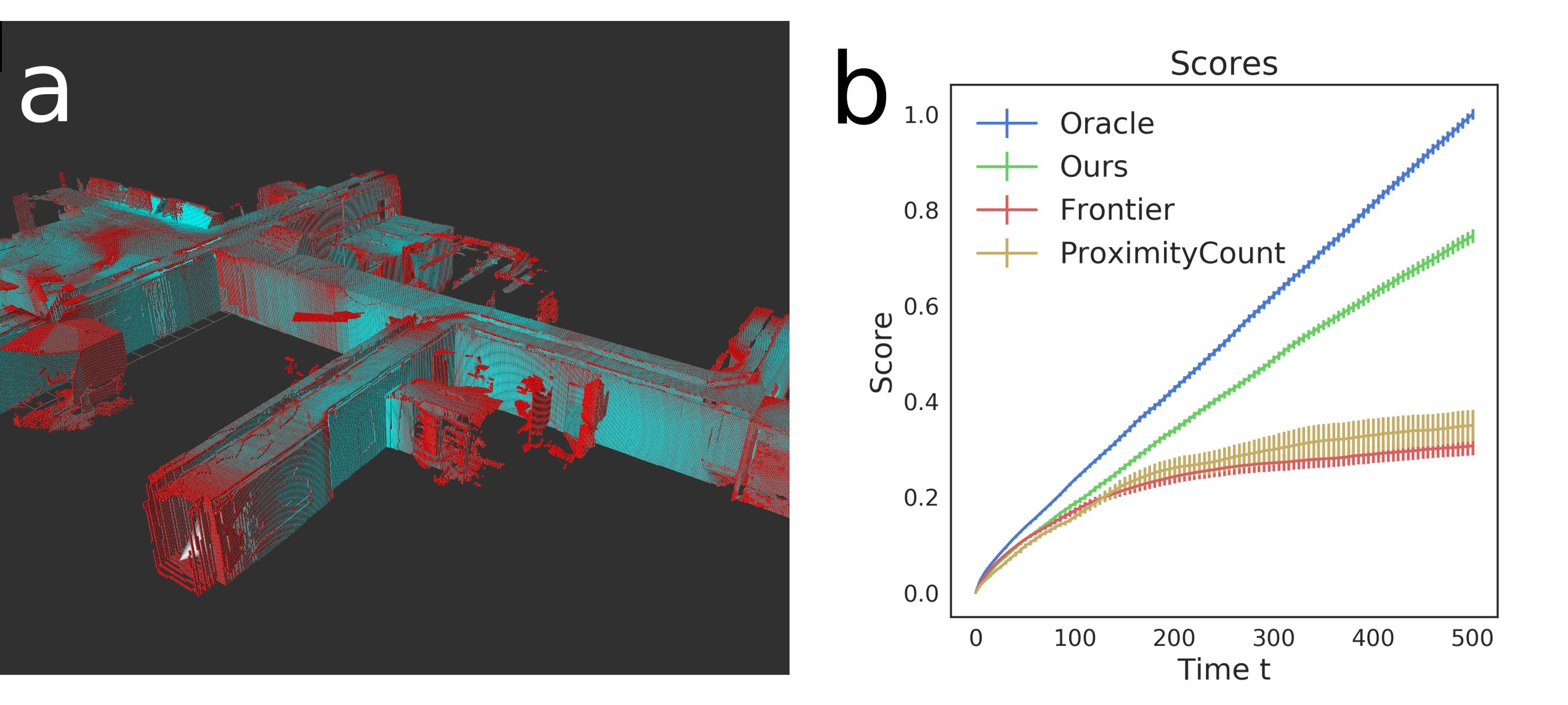}
	\caption{
		Shown are example explorations on \emph{real} indoor data --
		(a) Occupancy map of S3Dis Area5b (400 steps).
		(b) Performance plot for S3Dis Area5b (training on Area1).
		Color coding of observed voxels: High uncertainty (red) and low uncertainty (cyan).
	}
	\label{fig:additional_results_indoor}
\end{figure}

To show that our method is general and also works with real scenes we conducted additional experiments
on high-fidelity 3D reconstructions of buildings and on the 2D-3D-S indoor dataset \cite{armeni2017joint}
that was acquired with a Matterport\footnote{\url{https://matterport.com/}} camera.
Result are shown in Tab.\ \ref{tab:additional_results},
Fig.\ \ref{fig:additional_results_outdoor} and Fig.\ \ref{fig:additional_results_indoor}.
For the outdoor case we trained our model on the church (Fig.\ \ref{fig:additional_results_outdoor}a)
and evaluated on the historic building (Fig.\ \ref{fig:additional_results_outdoor}.c).
Despite the differences of both buildings in terms of geometry and scale
(the historic building is about $2x$ smaller in each dimension) our model is able to generalize.
For the indoor case we trained on \emph{Area1} and evaluated on \emph{Area5b} of the 2D-3D-S indoor dataset
\cite{armeni2017joint}.
Both experiments demonstrate that our method also works with real detailed scenes.

\begin{table}
	\centering
	\begin{tabular}{lcccc}
		\hline
		\multicolumn{5}{c}{Evaluation on additional real data} \\
		& \ Frontier\  & \ ProximityCount~\cite{isler2016information}\ 
		& \ \textbf{Ours}\  & \ {\color{gray}Oracle (GT access)}\  \\
		\hline
		Outdoor & $0.46$ & $0.58$ & $\mathbf{0.90}$ & \color{gray}$1.00$ \\
		Indoor & $0.44$ & $0.52$ & $\mathbf{0.78}$ & \color{gray}$1.00$ \\
		\hline
	\end{tabular}
	\caption{
		Comparison of \textit{Efficiency} metric on the additional real data.
		Our method achieves a higher value
		than the other utility functions on both ourdoor and indoor scenes.
		Note that in both cases the model was trained on data recorded from a single scene that was
		different from the evaluation scene.
		\textit{Efficiency} values are normalized with respect to the oracle for easier comparison.
	}
	\label{tab:additional_results}
\end{table}

\sectionvspace{-10pt}
\section{Discussion and Conclusions}

We presented an approach for efficient exploration of unknown 3D environments by predicting the utility of new views using a 3D ConvNet.
We input a novel multi-scale voxel representation of an underlying occupancy map, which represents the current model of the environment.
Pairs of input and target utility score are obtained from an oracle that has access to ground truth information.
Importantly, our model is able to generalize to scenes other than the training data and the underlying occupancy map enables robustness to
noisy sensor input such as depth images from a stereo camera. Experiments indicate that our approach improves upon previous methods in terms of reconstruction efficiency.

Limitations of our method include dependence on the surface voxel distribution
in the training data. In future work, it would be interesting to see how the method performs
on vastly different geometries such as rock formations and other
natural landscapes.
Similarly, our model is bound to the map resolution and mapping parameters
used in the training data.

Another limitation is the underlying assumption on a static scene. A dynamic object such as a human walking
in front of the camera would lead to occupied voxels that do not correspond to a static object.
While these voxels can change their state from occupied to free after additional observations if the human
walked away the intermediate occupancy map can lead to utility predictions that are not desired.
A possible solution to address this problem is to identify and segment dynamic objects in the depth maps
before integrating them into the occupancy map.

Our work suggests several directions for future work.  We used our learned utility function to implement a greedy next-best-view algorithm; however, our utility function could be used to develop more sophisticated policies that look multiple steps ahead.  In addition, our approach could be extended to be used in a generative way to predict future states of the 3D occupancy map or to predict 2D depth maps for unobserved views.  This could be used for model completion or hole-filling
which has numerous applications in computer vision and robotics.




\bibliographystyle{splncs04}
\bibliography{references}

\end{document}